\documentclass[sigconf]{acmart}

\usepackage{soul}
\usepackage{hyperref} 
\usepackage{algorithm}
\usepackage{algorithmic}
\usepackage{amsfonts}
\usepackage{stfloats}
\usepackage{multirow}
\usepackage{xcolor}
\usepackage{bm}
\usepackage{listings}
\usepackage{natbib}
\usepackage{enumitem}

\AtBeginDocument{%
  \providecommand\BibTeX{{%
    \normalfont B\kern-0.5em{\scshape i\kern-0.25em b}\kern-0.8em\TeX}}}

\copyrightyear{2023}
\acmYear{2023}
\setcopyright{acmlicensed}\acmConference[WWW '23]{Proceedings of the ACM Web Conference 2023}{April 30-May 4, 2023}{Austin, TX, USA}
\acmBooktitle{Proceedings of the ACM Web Conference 2023 (WWW '23), April 30-May 4, 2023, Austin, TX, USA}
\acmPrice{15.00}
\acmDOI{10.1145/3543507.3583285}
\acmISBN{978-1-4503-9416-1/23/04}

\begin{document}
\title{CTRLStruct: Dialogue Structure Learning for Open-Domain Response Generation}

\author{Congchi Yin}
\affiliation{%
  \institution{Nanjing University of Aeronautics and Astronautics}
  \city{Nanjing}
  \country{China}}
\email{congchiyin@nuaa.edu.cn}

\author{Piji Li}
\authornote{Corresponding author.}
\affiliation{%
  \institution{Nanjing University of Aeronautics and Astronautics}
  \city{Nanjing}
  \country{China}}
\email{pjli@nuaa.edu.cn}

\author{Zhaochun Ren}
\affiliation{%
  \institution{Shandong University}
  \city{Qingdao}
  \country{China}}
\email{zhaochun.ren@sdu.edu.cn}

\begin{abstract}
Dialogue structure discovery is essential in dialogue generation. Well-structured topic flow can leverage background information and predict future topics to help generate controllable and explainable responses. However, most previous work focused on dialogue structure learning in task-oriented dialogue other than open-domain dialogue which is more complicated and challenging. In this paper, we present a new framework \textbf{CTRLStruct} for dialogue structure learning to effectively explore topic-level dialogue clusters as well as their transitions with unlabelled information. Precisely, dialogue utterances encoded by bi-directional Transformer are further trained through a special designed contrastive learning task to improve representation. Then we perform clustering to utterance-level representations and form topic-level clusters that can be considered as vertices in dialogue structure graph. The edges in the graph indicating transition probability between vertices are calculated by mimicking expert behavior in datasets. Finally, dialogue structure graph is integrated into dialogue model to perform controlled response generation. Experiments on two popular open-domain dialogue datasets show our model can generate more coherent responses compared to some excellent dialogue models, as well as outperform some typical sentence embedding methods in dialogue utterance representation. Code is available in GitHub\footnote{\url{https://github.com/lemonsis/CTRLStruct}}.
\end{abstract}

\ccsdesc[500]{Computing methodologies~Discourse, dialogue and pragmatics}
\ccsdesc[500]{Computing methodologies~Natural language generation}

\keywords{Dialogue Structure Learning, Open-Domain Dialogue Generation, Utterance Representation, Contrastive Learning, Imitation Learning}

\maketitle
\section{Introduction}
\begin{figure}
\centering
\includegraphics[width=0.46\textwidth]{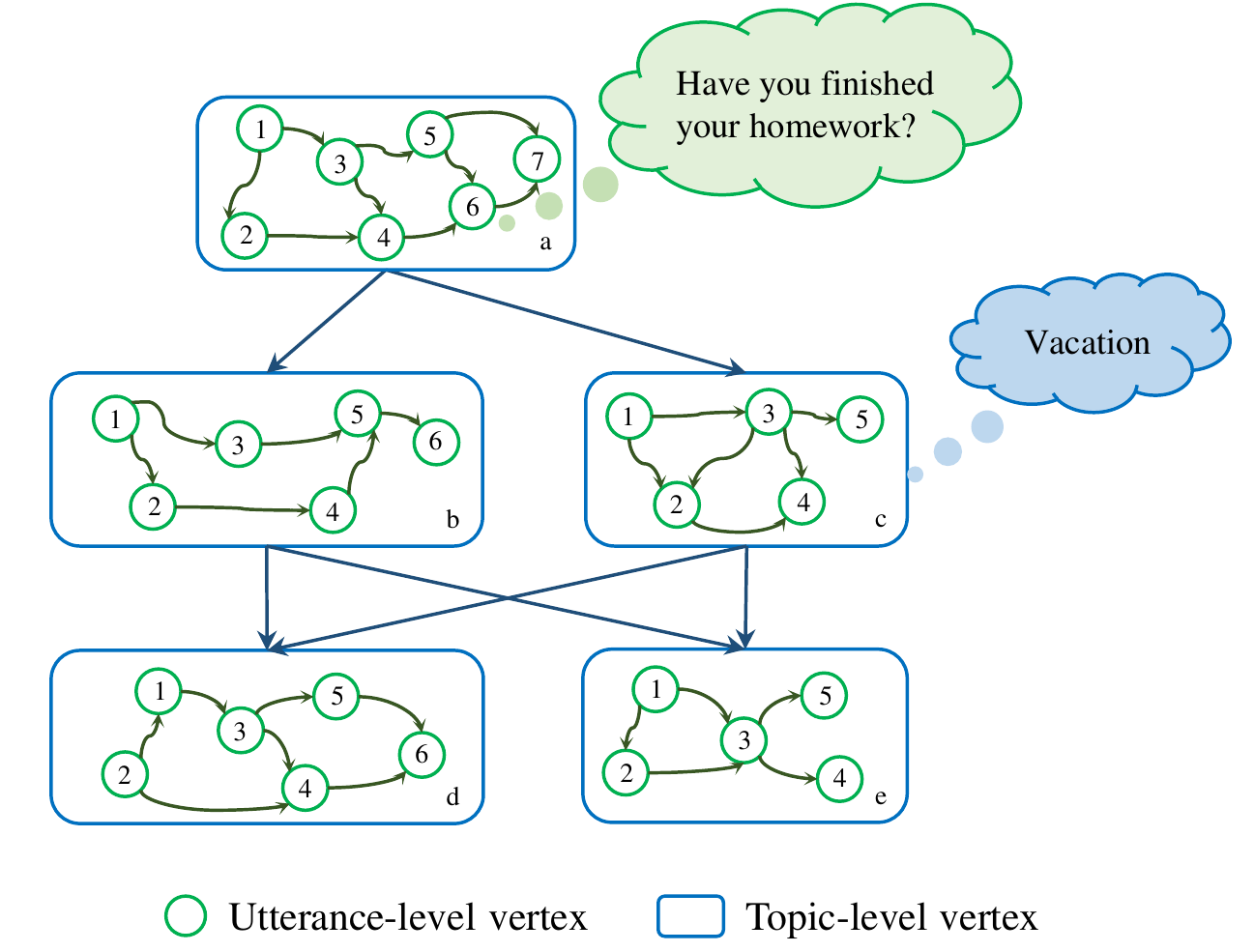}
\caption{Diagram of dialogue structure where circle stands for utterance-level structure and rounded rectangle stands for topic-level structure. The arrows indicate transitions.}
\label{f11}
\Description[Diagram showing what is dialogue structure]{Dialogue structure indicates utterance-level transition and topic-level transition.}
\end{figure}

Open-domain dialogue generation is a challenging task due to its complex multi-turn interacting structure, diverse topics, and lack of specific goals. As a result, only considering language modeling in response generation is far from enough like past work, such as GPT2 \cite{radford2019language}, BART \cite{DBLP:conf/acl/LewisLGGMLSZ20}, and T5 \cite{DBLP:journals/jmlr/RaffelSRLNMZLL20}. Dialogue structure, indicating dialogue states and their transitions, plays an integral part in dialogue generation. It's typically regarded as the combination of utterance-level structure and topic-level structure \cite{DBLP:conf/aaai/XuWNWC20}. The former can be viewed as flow of dialogue utterances while the latter stands for the topic transitions. As illustrated in Figure~\ref{f11}, transition from ``Have you finished you homework'' to ``Yes, I have.'' is considered as utterance-level structure, while the topic-level structure is transitions from a higher perspective, like from ``study'' to ``vacation''. Intuitively, if dialogue structure is modeled in multi-turn conversations, the chatbot can realize which topic it is in and what topic might be involved in the next step, which will largely improve multi-turn coherence as well as controllability and interpretability.

Dialogue structure was first mentioned in task-oriented dialogue researches. Early researches relied on hand-crafted rules \cite{DBLP:journals/nle/LarssonT00} or human annotations \cite{jurafsky1997switchboard} to model the structure in domain specific tasks. With the development of unsupervised learning, approach like Hidden Markov Model (HMM) shows promise in modeling latent structure of dialogues \cite{chotimongkol2008learning,DBLP:conf/acl/ZhaiW14,ritter-etal-2010-unsupervised}. Some of these studies are successfully applied in designing tour guide robot or ticket booking system. However it is difficult to reproduce the success when it comes to open-domain dialogue systems. Unlike task-oriented dialogues that only involve dozens or hundreds of dialogue states and simple state transitions, open-domain dialogue structure is rather difficult to model because of its unlimited number of dialogue states with complex and unstable flows. As people are likely to respond differently to the same utterance or topic in real-life scenarios. Recent studies \cite{wu-etal-2019-proactive,DBLP:conf/aaai/XuWNWC20} highlight the significance of building a topic-aware dialogue system to conduct proactive conversations. \citet{DBLP:conf/acl/0027L0NWC20} explicitly model the dialogue structure in open-domain dialogue corpus, and they combined variational auto-encoder and graph neural network to build utterance-level and session (topic)-level dialogue structure graph. However, the transition probability of dialogue states is calculated manually through co-occurrence frequency statistics, which is inefficient via offline computing and also needs some tuning work for the thresholds to guarantee the performance. Therefore, automatic dialogue structure extraction and modeling from dialogue corpus via appropriate machine learning approaches are required especially in the current training scenarios with easily obtained large-scale datasets.

Moreover, during the investigations \cite{https://doi.org/10.48550/arxiv.2203.03047, https://doi.org/10.48550/arxiv.2204.12749, DBLP:conf/naacl/ShiZY19}, we notice that the performance of dialogue utterance representation exerts huge impact on the quality of the learned dialogue structure. To improve the utterance representation learning especially in the dialogue scenarios, the unique dialogue-sensitive characteristics and features compared to other forms of corpora need to be deeply considered. Human dialogue can be viewed as a directional flow of topics, which is highly contextual and sequential, where ``contextual'' means that one response should be closely related to its context in topic level, while ``sequential'' means dialogue is composed of sequential utterances with internal logic that cannot be casually changed. Previous work \cite{DBLP:conf/acl/0027L0NWC20,DBLP:conf/ijcai/00080XXSL22} applied RNN or BERT \cite{DBLP:conf/naacl/DevlinCLT19} to directly encode and represent the utterance semantic information. These methods regard dialogue as ordinary corpus and neglect its unique dialogue-sensitive features, thus cannot guarantee high quality performance in our dialogue structure learning requirements.

To address the aforementioned issues, we present a new framework named \textbf{CTRLStruct}, including \ul{C}on\ul{T}rastive utterance \ul{R}epre-\\
sentation \ul{L}earning and Dialogue \ul{Struct}ure Modeling, to effectively extract utterance-level and topic-level transitions from large unlabelled data and generate coherent dialogues. In utterance representation learning, we design a special loss function via contrastive learning with dialogue-sensitive features considered, which can help generate dialogue utterance representation with rich semantics. Then utterance-level representations sharing similar meanings are gathered to form topic-level clusters, which are depicted as blue rounded rectangles in Figure~\ref{f11}. As both utterance-level and topic-level transitions are known in datasets, we apply imitation learning to mimic human intentions by calculating transitions between topic-level clusters to conduct dialogue structure learning and obtain the final dialogue structure graph. Once the dialogue structure graph is built, Kullback-Leibler divergence \cite{kullback1951information} between generated sentence and topic-level cluster center is considered to stimulate the model to generate topic-related conversations.  Experimental results on two open-domain dialogue datasets show that our model can generate more coherent and topic-related responses compared to some strong dialogue generation models, as well as outperform some strong sentence embedding methods in dialogue utterance representation.

The main contributions can be summarized as follows:
\begin{itemize}[leftmargin=*]
\item We propose a novel framework to discover dialogue structure and control response generation with unlabelled corpus. 
\item A contrastive learning encoder considering dialogue-sensitive characteristics is proposed to enhance utterance representation learning to further improve the performance of topic-level dialogue structure modeling.
\item We propose an imitation learning based method to conduct topic-level dialogue transition probability estimation and high-quality dialogue structure learning.
\item Both automatic and human evaluation show that dialogue structure helps promote dialogue generation model by generating topic-related and coherent conversations.
\end{itemize}

\section{Related Work}
\subsection{Dialogue Structure Learning}
Most previous unsupervised methods of discovering dialogue structure focused on task-oriented dialogues. Hidden Markov model \cite{DBLP:conf/naacl/RitterCD10,DBLP:conf/acl/ZhaiW14} is utilized to model dialogue flow on topic-level dialogue states. \citet{DBLP:conf/naacl/ShiZY19} adopted recurrent variational auto-encoder in modeling latent dialogue structure with highly non-linear dynamics. \citet{GUNASEKARA201917} built dialogue structure upon quantizing dialogue space into clusters. Latest researches \cite{DBLP:conf/aaai/SunST0DYSHS21,DBLP:conf/acl/0027L0NWC20,DBLP:conf/ijcai/00080XXSL22} applied graph neural network to further improve hierarchical dialogue structure modeling. Some of these work began to shed light on discovering dialogue structure in open-domain settings.

\begin{figure*}[t]
\centering
\includegraphics[width=\linewidth]{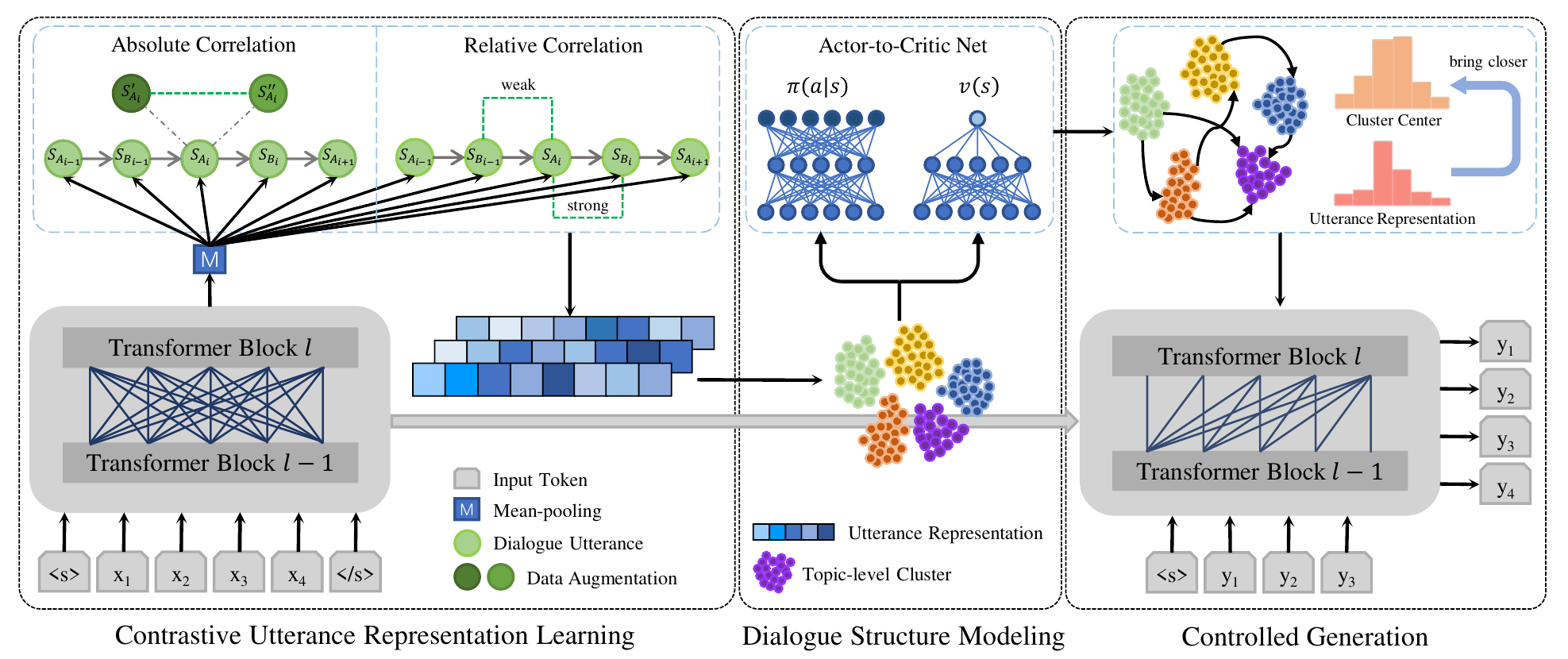}
\caption{The framework of the whole model. \textless$s$\textgreater and \textless/$s$\textgreater in the input are special tokens indicating begin of sentence and end of sentence respectively. Notice that the bi-directional attention encoder trained through contrastive learning in CTRLStruct is not used in dialogue generation, original pre-trained encoder is used instead.}
\Description[The main framework of CTRLStruct]{The framework of CTRLStruct consists of three main components: contrastive utterance representation learning part, dialogue structure modeling part and controlled generation part.}
\label{f1}
\end{figure*}

\subsection{Pre-trained Language Model}
Large language models pre-trained on massive corpora have promoted the development of various NLP tasks. They can be finetuned for different downstream tasks and show competitive performance. BERT \cite{DBLP:conf/naacl/DevlinCLT19} uses Transformers \cite{DBLP:conf/nips/VaswaniSPUJGKP17} based on bi-directional attention mechanism where context is considered in attention score calculation. GPT \cite{radford2018improving}, GPT2 \cite{radford2019language}, and GPT3 \cite{DBLP:conf/nips/BrownMRSKDNSSAA20} use unidirectional attention Transformers where one token is only allowed to attend its previous tokens. Recent researches \cite{DBLP:conf/acl/LewisLGGMLSZ20} also combined  bi-directional attention and unidirectional attention to create models good at both natural language understanding and natural language generation.
\subsection{Contrastive Learning}
Contrastive learning is a self-supervised framework in learning meaningful representations. Its main purpose is to bring similar samples closer and separate dissimilar samples. \citet{DBLP:journals/jmlr/GutmannH10} proposed Noise-Contrastive Estimation and \citet{DBLP:journals/corr/abs-1807-03748} developed it into InfoNCE loss, which is
\begin{equation}
\label{infonce}
    loss_i=-\log \frac{e^{sim(\bm{z}_i,\bm{z}_i^+)/ \tau}}{\sum_{j=1}^{N} e^{sim(\bm{z}_i,\bm{z}_j)/\tau}}.
\end{equation}
In Equation (\ref{infonce}), $\bm{z}_i$ and $\bm{z}_i^+$ are representations of positive pairs $\bm{x}_i$ and $\bm{x}_i^+$. $\tau$ is a temperature hyperparameter and $sim(\bm{z}_i,\bm{z}_i^+)$ indicates some distance measurement between $\bm{z}_i$ and $\bm{z}_i^+$.
InfoNCE loss lays solid foundation for modern contrastive learning methods like MoCo \cite{DBLP:conf/cvpr/He0WXG20} and SimCLR \cite{DBLP:conf/icml/ChenK0H20} in computer vision. In NLP domain, SimCSE \cite{DBLP:conf/emnlp/GaoYC21} uses dropout in Transformers to build positive samples and successfully learns state-of-the-art sentence embedding.

\section{Methodology}
\subsection{Overview}
Given a dialogue corpus $\mathcal{D}$ that contains $|D|$ conversations denoted as $\{X_1,X_2,\ldots,X_{|D|}\}$, each conversation is composed of multi-turn utterances of two speakers, $A$ and $B$. The $i$-th conversation $X_i$ contains $k_i$ utterances $X_i=[S_{A_1},S_{B_1},S_{A_2},\ldots]$. In dialogue structure modeling, utterances with similar semantic meanings are supposed to gather in the same topic cluster $C_i$. We assume that each cluster center vector $\bm{c}_i$ contains topic-level 
information. We take these clusters as random variables in Markov chain, where each has certain probability of moving to another. Our goal is to figure out the dialogue structure consisted of topic transitions, and utilize the dialogue structure to control response generation.

As shown in Figure \ref{f1}, our proposed CTRLStruct model consists of three main components: contrastive utterance representation learning part, dialogue structure modeling part, and dialogue structure controlled response generation part. In the first part, the mean-pooling output of a bi-directional Transformer block is viewed as original version of utterance representation. Then we apply contrastive learning to further train the Transformer encoder and get the final utterance representation. In the second part, we gather semantically similar utterances together to form topic-level clusters and utilize imitation learning to automatically calculate transition probabilities among topics, and then the dialogue structure graph has been built. To conduct dialogue structure controlled response generation, given the dialogue context, we can obtain the target topic according to the dialogue structure graph. Then during training, we bring the representation of generated response closer to the predicted cluster center. In this way topic-level dialogue structure information is integrated in the training procedure of the auto-regressive decoder.
 
\subsection{Utterance Representation Learning}
For each input utterance $S$, the bi-directional Transformer encoder outputs original representation as  $\mathbf{H} \in \mathbb{R}^{m \times n}$ where $m$ stands for the number of tokens in utterance. Previous work \cite{DBLP:conf/emnlp/ReimersG19,DBLP:conf/emnlp/LiZHWYL20} showed that taking mean-pooling of pre-trained models' output embeddings leads to better performance than max-pooling or $[CLS]$ representation. We perform mean-pooling to $\mathbf{H}$ and get $n$-dimensional vector $\bm{h}$ as preliminary utterance representation. 

Different from other forms of corpus, dialogue has its unique features that are non-negligible in dialogue utterance representation learning. It can be viewed as directional flow of topics from a higher perspective. So we make the basic assumption that human dialogue is highly \textbf{contextual} and \textbf{sequential}. ``Contextual'' means that one response should be closely related to its context in topic level. The closer one utterance is to another, the more relevant it is to that utterance. ``Sequential'' means dialogue is composed of sequential utterances with internal logic. Conversations like ``A: Are you free to climb mountain with us tomorrow? B: Sorry, I have to finish my job first.'' are conventional, but if its order is changed, that would be weird. However, even with the defined assumptions above, it is still hard to develop a set of rules to clearly explain the role one utterance plays in the dialogue. One utterance might be semantically similar with its previous utterance, or has few connection with the previous one but has a close relationship with its next response. For example, when an utterance is a response to some queries, it satisfies the former situation. When an utterance is the beginning of a new topic, it corresponds to the latter situation. We define this problem as Utterance Ascription Issue.

We apply contrastive learning to handle the above-mentioned problem. Two kinds of correlation are defined: \textbf{Absolute Correlation} and \textbf{Relative Correlation}. As is illustrated in Figure \ref{f1}, Absolute Correlation follows the contrastive learning framework proposed by SimCLR \cite{DBLP:conf/icml/ChenK0H20}. Two data augmentation samples $S_{A_i}^{\prime}$ and $S_{A_i}^{\prime\prime}$ of one utterance $S_{A_i}$ constitute a positive pair. Specifically, we choose four different data augmentation strategies in CTRLStruct, which are insert contextual relevant words, random replacement, synonym substitution, and dropout augmentation trick applied in SimCSE \cite{DBLP:conf/emnlp/GaoYC21}. Relative Correlation is composed of \textbf{Strong Relativity} and \textbf{Weak Relativity}. Considering the sequential feature of dialogue, the relativity between $S_{A_i},S_{B_i}$ is not equal to that between $S_{B_i},S_{A_i}$ like sort of non-metric distance measurement. Take three sequential utterances $S_{B_{i-1}},S_{A_i},S_{B_i}$ in dialogue for example. As topics flow directionally in dialogue, we define Strong Relativity for $S_{A_i}$ as $S_{A_i}$ and its next response $S_{B_i}$ making up a positive pair. Weak Relativity stands for $S_{A_i}$ and its previous response $S_{B_{i-1}}$ making up a weak positive pair. Strong Relativity and Weak Relativity are critical in solving the Utterance Ascription Issue. When an utterance is semantically similar with its previous utterance, the constraint of Weak Relativity can maintain the relationship so as not to be ruined by Strong Relativity. When an utterance is similar with its next utterance, Strong Relativity can bring them closer in semantic space. The relations are vividly shown in Figure \ref{f1}.

Under the above settings, we design Absolute Correlation Loss $l_{AC}$ and Relative Correlation Loss $l_{RC}$. Supposing $S_{A_i}^{\prime}$ and $S_{A_i}^{\prime\prime}$ are two data augmentation utterances of $S_{A_i}$, $\bm{h}$ is the original encoder representation of utterance $S$. Absolute Correlation Loss of response $S_{A_i}$ is defined as
\begin{equation}
    \label{equ1}
    l_{AC}(S_{A_i}^{\prime},S_{A_i}^{\prime\prime}) = -\log \frac{e^{sim(\bm{h}_{A_i}^{\prime},\bm{h}_{A_i}^{\prime\prime})/\tau}}{\sum_{S_i \in \bigcup_{j=1}^{2|D|} X_j} \mathbf{1}_{[i \neq A_i]} e^{sim(\bm{h}_{A_i}^{\prime},\bm{h}_i)/\tau}},
\end{equation}
where $\mathbf{1}_{[i \neq A_i]}\in\{0,1\}$ is an indicator function evaluating to $1$ iff $i \neq A_i$, $sim(\bm{a},\bm{b})$ is the cosine similarity $\frac{\bm{a}^ \top \bm{b}}{\|\bm{a}\| \|\bm{b}\|}$ between vector $\bm{a}$ and $\bm{b}$, $X_j$ is the $j$-th conversation. The capacity of set is $2|D|$ because we use one utterance's two augmentation samples for contrastive learning. Equation (\ref{equ1}) also applies to speaker $B$'s responses. Relative Correlation Loss consists of two losses, Strong Relativity Loss $l_{SR}$ and Weak Relativity Loss $l_{WR}$. Strong Relativity Loss is
\begin{equation}
\label{equ3}
    l_{SR}(S_{A_i},S_{B_i}) = -\log \frac{e^{sim(\bm{h}_{A_i},\bm{h}_{B_i})/\tau}}{\sum_{S_i \in \bigcup_{j=1}^{|D|} X_j} \mathbf{1}_{[i \neq A_i]} e^{sim(\bm{h}_{A_i},\bm{h}_i)/\tau}},
\end{equation}
where $S_{B_i}$ is the next sentence of $S_{A_i}$. Similarly, Weak Relativity Loss is defined as $S_{A_i}$ and its previous response,
\begin{equation}
    \label{equ4}
    l_{WR}(S_{A_i},S_{B_{i-1}}) = - \log \frac{\lambda_1 e^{sim(\bm{h}_{A_i},\bm{h}_{B_{i-1}})/\tau}}{\sum_{S_i \in \bigcup_{j=1}^{|D|} X_j} \mathbf{1}_{[i \neq A_i]}e^{sim(\bm{h}_{A_i},\bm{h}_i)/\tau}}.
\end{equation}
Compared to $l_{SR}$, Weak Relativity is reflected in the coefficient $\lambda_1$. The total Relative Correlation Loss is the sum of Strong Relativity Loss and Weak Relativity Loss
\begin{equation}
\label{equ5}
    l_{RC} = l_{SR}+l_{WR}.
\end{equation}
Since mini-batch gradient descent is used to optimize neural network, Absolute Correlation Loss and Relative Correlation Loss of batch with size $N$ is written as
\begin{equation}
\label{equ6}
    Loss_{AC} = \frac{1}{2N} \sum_{i=1}^{N} [l_{AC}(S_{i}^{\prime},S_{i}^{\prime\prime})+l_{AC}(S_{i}^{\prime\prime},S_{i}^{\prime})],
\end{equation}
\begin{equation}
    \label{equ7}
    Loss_{RC} = \frac{1}{N-1} [\sum_{i=1}^{N-1} l_{SR}(S_i,S_{i+1}) + \sum_{i=2}^{N} l_{WR}(S_i,S_{i-1})].
\end{equation}
The total loss in utterance representation training is the sum of Absolute Correlation Loss and Relative Correlation Loss
\begin{equation}
    \label{equ8}
    Loss_{total} = Loss_{AC}+Loss_{RC}.
\end{equation}
The whole self-supervised training process of contrastive utterance representation is conducted on bi-directional Transformer encoder. When it finishes, the encoder outputs utterance representation for dialogue structure modeling in next part.

\subsection{Dialogue Structure Modeling}
We perform clustering to utterance representations, aiming to gather utterances sharing similar topics. We utilize K-Means with cosine similarity as clustering methods and achieve good clustering performance and conciseness.

Utterance representations are gathered into $k$ topic-level clusters $C_1,C_2,\ldots,C_k$ whose center vectors are $\bm{c}_1,\bm{c}_2,\ldots,\bm{c}_k$ respectively. As shown in Figure \ref{f1}, our goal is to model dialogue structure graph $\mathcal{G}=\{\mathcal{V}, \mathcal{E}\}$, where topic-level clusters are taken as vertices and transition probabilities to other vertices are viewed as edges. In CTRLStruct, we don't explicitly model the utterance transitions inside certain topic cluster like other methods. Because Large pre-trained language model is capable of fixing the problem. With such topic-level dialogue structure graph, our model can predict whether to continue talks related to current topic or transit to another topic during conversations.

Imitation learning \cite{DBLP:journals/csur/HusseinGEJ17, DBLP:journals/ras/ArgallCVB09} is applied in transition probability calculation. However, environment or simulator is not available in our faced problem. We only have dialogue datasets with unknown topic state transition probabilities, which can be viewed as offline environment in reinforcement learning concept. Fortunately, \citet{DBLP:conf/nips/RajaramanYJR20} prove the optimality of behavioral cloning under the settings that the state transition function is unknown and the environment for simulation is unavailable. The idea of behavioral cloning \cite{DBLP:conf/mi/BainS95, DBLP:journals/neco/Pomerleau91} is attempt to recover expert policy with given high-quality dataset. Since the ultimate goal of open-domain dialogue system is to achieve human-level conversation quality, we apply behavioral cloning to mimic topic transitions from expert trajectories in high-quality dialogue datasets. Utterance representation $bmh_i$ is taken as state and cluster center vector $\bm{c}_j$ is considered as action in behavioral cloning setting. The size of state space and action space are $\mathcal{S} \in \mathbb{R}^{u \times n}$ and $\mathcal{A} \in \mathbb{R}^{k \times n}$ respectively, where $u$ is the total number of utterances and $k$ is the number of clusters. At time step $t$, action $\bm{c}_{t+1}$, which is the center vector of state $\bm{h}_{t+1}$'s cluster, will be taken and state $\bm{h}_t$ is to transit to state $\bm{h}_{t+1}$. So the expert trajectory can be written as $\bm{h}_1,\bm{c}_2,\bm{h}_2,\bm{c}_3,\bm{h}_3,\ldots,\bm{h}_m$.

From the above settings, state space $\mathcal{S}$ can be considered as non-discrete space for $u$ is too large while action space $\mathcal{A}$ is discrete space. However in real training process we view action space as continuous space for the diversity of topics (actions). We select the cluster center vector that is closest to calculated action in cosine similarity as final action to take in current state. Under such situation in behaviour cloning we usually use Maximum Likelihood Estimation (MLE) to directly estimate the policy $\widehat{\pi}_\theta$. Since $\mathcal{S},\mathcal{A} \in \mathbb{R}$, the optimization objective can be written as
\begin{equation}
    \label{equ9}
\max _{\theta} \sum_{(\bm{h}, \bm{c}) \in \mathcal{D}} \log \left(\widehat{\pi}_{\theta}(\bm{c} \mid \bm{h})\right) .
\end{equation}
Gaussian distribution is adopted to represent the policy as most behavioral cloning methods do \cite{DBLP:journals/jmlr/RossB10} in continuous action or state space. For each state $\bm{h}$, we assume policy $\widehat{\pi}_{\theta}(\cdot| \bm{h})\sim\mathcal{N}\left(\mu_{\theta}(\bm{h}), \sigma_{\theta}^{2}(\bm{h})\right)$ where $\mu_{\theta}(\bm{h})$ and $\sigma_{\theta}^{2}(\bm{h})$ are the mean and variance
\begin{equation}
    \label{equ9.5}
    \widehat{\pi}_{\theta}(\cdot| \bm{h})=\frac{1}{\sqrt{2\pi}\sigma_{\theta}(\bm{h})}e^{-\frac{(c-\mu_{\theta}(\bm{h}))^2}{2{\sigma_{\theta}(\bm{h})}^2}}.
\end{equation}
According to Equation (\ref{equ9.5}), Equation (\ref{equ9}) can be reduced to
\begin{equation}
    \label{equ10}
    \min _{\theta} \sum_{(\bm{h}, \bm{c}) \in \mathcal{D}} \frac{\left(\bm{c}-\mu_{\theta}(\bm{h})\right)^{2}}{2 \sigma_{\theta}^{2}(\bm{h})}+\frac{1}{2} \log \left(2 \pi \sigma_{\theta}^{2}(\bm{h})\right).
\end{equation}
We use Actor-to-Critic network \cite{DBLP:journals/tnn/SuttonB98} to estimate the value of $\mu_{\theta}(\bm{h})$. Specifically, both Actor network and Critic network are consisted of fully connected layers. Critic network estimates value through state and Actor network figures out what action to take under certain state. Since reward is not available in behaviour cloning, we don't consider it in Actor-to-Critic network. Variance $\sigma_{\theta}^{2}(\bm{h})$ is set as constant independent of parameter $\theta$. So the final object to optimize is translated into mean square error regression problem
\begin{equation}
\label{equ11}
\min _{\theta} \sum_{(\bm{h}, \bm{c}) \in \mathcal{D}}\left(\bm{c}-\mu_{\theta}(\bm{h})\right)^{2}.
\end{equation}
After solving the regression problem with neural network, what action agent at certain state will take as well as the corresponding probability can be predicted by the policy $\pi_\theta$.

\subsection{Dialogue Structure Controlled Generation}
In the dialogue generation stage, we follow previous work \cite{radford2018improving} to factorize the joint probabilities over tokens as the product of conditional probabilities
\begin{equation}
    \label{equ12}
    p(Y)=\prod_{i=1}^{n} P\left(t_{i} \mid t_{1},\ldots, t_{i-1}\right),
\end{equation}
where $Y$ is one response that contains a sequence of tokens $t_1,t_2,\ldots$
$,t_n$. Auto-regressive model relies on Transformer decoder blocks with left-to-right attention to generate tokens one after another. In our encoder-decoder model CTRLStruct, dialogue history $X_i$ is sent to encoder and the decoder generates responses in auto-regressive way. We optimize the negative log likelihood of original response:
\begin{equation}
    \label{equ13}
    l_{NLL}(Y) = -\sum_{i=1}^{n}\log P(t_i \mid t_{1},\ldots, t_{i-1}).
\end{equation}

Dialogue structure graph is integrated into the auto-regressive Transformer decoder in the following manner. What cluster $C_{i+1}$ the next response will be in through $\pi_{\theta}(\cdot| \bm{h}_{i})$ can be predicted. So when the auto-regressive model finishes outputting tokens, we take the mean-pooling of model's last hidden state output to get the representation $\bm{h}_{i+1}$ of the generated utterance. The representation vector and cluster center vector $\bm{c}_{i+1}$ can be viewed as one dimensional probability distribution as well, representing utterance distribution in semantics space and topic space respectively. We adopt Kullback-Leibler divergence \cite{kullback1951information} to bring distribution in semantic space closer to topic space, which ensures the generated utterance's relevance to its supposed topic. Kullback-Leibler divergence can be written as
\begin{equation}
    \label{equ14}
    D_{KL}(\bm{h}||\bm{c})=\sum_{x \in X}[\bm{h}_{i+1}(x) \log \frac{\bm{h}_{i+1}(x)}{\bm{c}_{i+1}(x)}].
\end{equation}
 The total loss to optimize in the decoder is the sum of negative log likelihood loss and Kullback-Leibler divergence
\begin{equation}
    \label{equ15}
    Loss_{Gen}=l_{NLL} + \lambda_{2} D_{KL}.
\end{equation}
Then the training process of CTRLStruct has finished. We can apply the encoder-decoder model for dialogue generation like other Transformer based generative language models.

\section{Experimental Setups}
\subsection{Research Questions}
We aim to answer the following research questions through experiments on response generation and utterance representation:
\begin{itemize}[leftmargin=*]
\item \textbf{RQ1}: How does CTRLStruct perform in open-domain dialogue compared to several strong dialogue generation models? 
\item \textbf{RQ2}: Does CTRLStruct really control the topic of response? Does the discovered dialogue structure help in response generation?
\item \textbf{RQ3}: How does CTRLStruct perform in dialogue utterance representation compared to other sentence embedding methods? Can semantically similar utterances cluster in CTRLStruct?
\item \textbf{RQ4}: How is the generalization ability of CTRLStruct? Can it be applied to models with other types of backbone?
\end{itemize}

\subsection{Datasets}
To evaluate the performance of dialogue structure learning in response generation, we conduct experiments on two popular open-domain dialogue datasets, DailyDialog \cite{DBLP:conf/ijcnlp/LiSSLCN17} and PersonaChat \cite{DBLP:conf/acl/KielaWZDUS18}. DailyDialog is a human-written multi-turn dialogue dataset which is less noisy and covers various topics about daily life. We use the dataset randomly separated by the authors as training/validation/test sets with 11,118/1,000/1,000 conversations. In PersonaChat chit-chat agent is endowed with a configurable and consistent persona to generate more personal, specific and engaging conversions. The original version of PersonaChat is divided as 10,907/1,000/968 dialogues for training/validation/test sets. In our work, conversations in DailyDialog and PersonaChat without any extra label or persona information are used during model training.

\subsection{Compared Methods}
CTRLStruct is compared to several strong models with respect to the performance of dialogue generation and utterance representation. To evaluate the quality of dialogue generation, traditional models and large pre-trained language models are selected:
\begin{itemize}[leftmargin=*]
\item \textbf{Seq2Seq}: Vanilla Sequence-to-Sequence model used in machine translation with attention mechanism \cite{DBLP:conf/nips/VaswaniSPUJGKP17}.
\item \textbf{CVAE}: CVAE \cite{DBLP:conf/acl/ZhaoZE17} is a generative model combining variational auto-encoder \cite{DBLP:journals/corr/KingmaW13} with RNN Sequence-to-Sequence framework.
\item \textbf{BART}: BART \cite{DBLP:conf/acl/LewisLGGMLSZ20} pre-trains a Sequence-to-Sequence model combining bi-directional and auto-regressive Transformers. BART-large is used in experiments.
\item \textbf{DialoGPT}: DialoGPT \cite{DBLP:conf/acl/ZhangSGCBGGLD20} extends GPT-2 \cite{radford2019language} to address conversational neural response generation by training on Reddit data. The medium version of DialoGPT is applied in experiments.
\item \textbf{BlenderBot}: BlenderBot \cite{DBLP:journals/corr/abs-2004-13637} which features Blended Skill Talk set-up is one of the most performing open-domain chatbots. We apply the 2.7B parameter model in our experiments.
\end{itemize}

To evaluate the utterance representation capability, we choose the following methods:
\begin{itemize}[leftmargin=*]
\item \textbf{BERT}: BERT \cite{DBLP:conf/naacl/DevlinCLT19} is composed of bi-directional Transformer encoders pre-trained by masked language modeling on next sentence prediction task.
\item \textbf{Unsupervised SimCSE}: Unsupervised version of SimCSE \cite{DBLP:conf/emnlp/GaoYC21} regards two augmented samples of one utterance as positive pair and perform contrastive learning, which is the state-of-the-art sentence embedding framework.
\end{itemize}

\begin{table*}
  \centering
  \caption{Automatic and human evaluations of our model and five other strong dialogue generation models on two open-domain dialogue datasets without additional labelled information. The highest values are written in bold.}
  \resizebox{\textwidth}{!}
  {
    \begin{tabular}{ccccccccc}
    \toprule
    \multirow{2}{*}{\textbf{Dataset}} & \multirow{2}{*}{\textbf{Model}} & \multicolumn{3}{c}{\textbf{Automatic Evaluation}} & \multicolumn{4}{c}{\textbf{Human Evaluation}} \\
       \cmidrule(r){3-5}  \cmidrule(r){6-9}&      & \multicolumn{1}{c}{\textbf{BLEU-1/2}} & \multicolumn{1}{c}{\textbf{Distinct-1/2}} & \multicolumn{1}{c}{\textbf{ROUGE-L}} & \multicolumn{1}{c}{\textbf{Fluency}} & \multicolumn{1}{l}{\textbf{Coherence}} & \multicolumn{1}{c}{\textbf{Informativeness}} & \multicolumn{1}{c}{\textbf{Overall}} \\
    \hline
    \multirow{8}{*}{PersonaChat} & Seq2Seq &   0.170 / 0.023    & 0.008 / 0.032   &   0.062    &    1.72   &    0.40   &    1.24   &  0.98\\        
& CVAE  &   0.192 / 0.041 &   0.018 / 0.073    &   0.059   &    1.76   &   0.64    &    1.36   &  1.28\\ 
& BART  &    0.291 / 0.109   &   0.030 / 0.107    &   0.154    &   1.84    &    0.82   &   1.42    &  1.38\\
&DialoGPT    & \textbf{0.324} / 0.114 &  0.031 / 0.137  &  0.188  &  1.84 & 1.04 & \textbf{1.80} & 1.58\\
& BlenderBot  & 0.280 / 0.112    &   \textbf{0.036} / \textbf{0.167}    &    \textbf{0.198}     &  \textbf{1.88}      &   1.20      &    1.76   &    1.52      \\

& CTRLStruct w/o Total\_Loss  &    0.267 / 0.092   &   0.024 / 0.077   &    0.151   &   1.80    &   0.76    &   1.54    &  1.26\\     
& CTRLStruct w/o WR\_Loss  &    0.306 / 0.113   &   0.030 / 0.111    &   0.159    &    1.86   &    1.28   &   1.70    &  1.56\\       
& CTRLStruct &   0.316 / \textbf{0.119}    &   0.032 / 0.114    &   0.161   &    \textbf{1.88}   &    \textbf{1.40}   &    1.72   &  \textbf{1.62}\\
    \midrule
    \multirow{8}{*}{DailyDialog} & Seq2Seq &   0.309 / 0.036   &   0.068 / 0.220    &   0.049    &   1.86    &   0.46    &   0.62    &  0.94\\
& CVAE  &   0.381 / 0.138    &   0.072 / 0.299   &    0.046   &    1.86   &   0.58    &    0.72   &  1.18\\       
& BART  & 0.364 / 0.141      &   0.112 / 0.378    &  0.075    &   1.92    &    1.66   &   1.38    &  1.54\\
& DialoGPT  &  0.353 / 0.134   &   0.106 / 0.352  &  0.105  &  1.90 & 1.60 & 1.46 & 1.60\\
& BlenderBot & 0.335 / 0.124  &  0.111 / 0.340   &  \textbf{0.108} &  1.94 & 1.64 &1.62 & 1.72\\

& CTRLStruct w/o Total\_Loss  &    0.346 / 0.130   &   0.095 / 0.328    &   0.076    &    1.90   &   1.34    &    1.24   &  1.36\\     
& CTRLStruct w/o WR\_Loss  &    0.388 / 0.151   &   0.112 / 0.387    &   0.077    &    1.92   &   \textbf{1.72}    &   1.56    &  1.70\\       
& CTRLStruct & \textbf{0.397} / \textbf{0.157}    &   \textbf{0.118} / \textbf{0.402}    &   0.080    &    \textbf{1.96}   &   \textbf{1.72} &   \textbf{1.68}    & \textbf{1.78} \\
    \bottomrule
    \end{tabular}
}
\label{tab:gen}
\end{table*}

\subsection{Evaluation Metrics}
We assess the performance of dialogue generation where both automatic and human evaluation metrics are applied. Automatic evaluation metrics include BLEU-1/2 \cite{DBLP:conf/acl/PapineniRWZ02}, Distinct-1/2 \cite{DBLP:conf/naacl/LiGBGD16} and ROUGE-L \cite{lin2004rouge}. BLEU, Distinct and ROUGE-L measure the n-gram overlap between generated utterance and ground truth, generation diversity and the number of longest common subsequence between generated utterance and ground truth respectively. In human evaluation, we follow settings in PLATO \cite{DBLP:conf/acl/BaoHWWW20}, where evaluators are supposed to score on a scale of $\{0,1,2\}$ from four aspects -- fluency, coherence, informativeness and overall. Zero indicates bad performance, one indicates normal and two stands for good responses. 

Moreover, we also analyze if the generated responses are truly related to its supposed topic. We design two evaluation metrics named Hard Topic Hit Accuracy (HTHA) and Soft Topic Hit Accuracy (STHA), aiming to prove CTRLStruct can control the topic flow in generation process compared to other models. HTHA is defined as the proportion of generated responses' topic clusters matching their pseudo-labels. Here pseudo-label stands for the cluster identity number of ground truth response, which is obtained through the dialogue structure modeling part in CTRLStruct. STHA is defined as the proportion of generated responses' topic clusters similar to their pseudo-labels. The similarity is measured through the cosine distance between generated response's cluster center vector and ground truth's cluster center vector. If the similarity exceeds given threshold $\varphi$, then the generated response is viewed as matching its pseudo-label. HTHA can be considered as special STHA whose threshold $\varphi=1.00$. Besides, macro-\emph{F}1 and micro-\emph{F}1 are also applied to evaluate model's topic-aware generation performance.

The evaluation of unsupervised utterance representation in dialogue is a challenge. We share the same sentiment with \citet{DBLP:conf/emnlp/ReimersG19} that utterance representation with good expressiveness can help semantically similar sentences cluster. Since ground truth labels of clusters are unknown, internal evaluation metrics including Calinski-Harabasz Index \cite{calinski1974dendrite} and Davies-Bouldin Index \cite{davies1979cluster} are used to assess the quality of clusters. Higher Calinski-Harabasz Index score and lower Davies-Bouldin Index score indicate better clusters definition and separation.

\subsection{Implemention Details}
In utterance representation training, we set the coefficient $\lambda_1$ of Weak Relativity Loss as $0.2$. We found that a large $\lambda_1$ will cause collapse in model training. Utterance representations are separated into $60$ clusters through K-Means. The model is trained for $20$ epochs with batch size of $256$. The temperature coefficient of models is set as $0.05$. We use nlpaug package \cite{ma2019nlpaug} for data augmentation.\footnote{\url{https://github.com/makcedward/nlpaug}}

In generation part, pre-trained BART-large is chosen as encoder and decoder in our model. Actor-to-Critic network is used in behavioral cloning in CTRLStruct. The coefficient $\lambda_2$ of KL-divergence is set as $1.2$ to reach the best performance in both datasets. In automatic evaluation, we use beam search in pre-trained language models while in human evaluation and topic matching evaluation, both top-$p$ sampling \cite{DBLP:conf/iclr/HoltzmanBDFC20} and top-$k$ sampling \cite{DBLP:conf/acl/LewisDF18} are adopted to generate more diverse conversations.

\begin{table*}[t]
  \centering
  \caption{Experimental results showing topic-level relatedness of generated responses and its pseudo-label. HTHA and STHA are short for Hard Topic Hit Accuracy and Soft Topic Hit Accuracy. The highest values are written in bold.}
    \resizebox{\textwidth}{!}
    {
    \begin{tabular}{ccccccccc}
    \toprule
    \textbf{Dataset} & \textbf{Model} & \textbf{macro-\emph{F}1} & \textbf{micro-\emph{F}1}   & \textbf{HTHA}  & \textbf{STHA ($\varphi=0.95$)} & \textbf{STHA ($\varphi=0.90$)} & \textbf{STHA ($\varphi=0.85$)} & \textbf{STHA ($\varphi=0.80$)} \\
    \midrule
    \multirow{8}{*}{PersonaChat} & Seq2Seq &   0.02    &   0.02    &   2.97\%    &   2.97\%    &   2.97\%    & 3.09\% & 5.75\%\\
          & CVAE  &    0.02   &    0.03   &   3.51\%    &   3.51\%    &   3.51\%    & 3.71\% & 7.20\%\\
          & BART  &    0.03   &  0.06     &    8.12\%   &    8.12\%   &    8.12\%   & 8.85\% & 18.00\% \\
          & DialoGPT &  0.03 &  0.05 & 7.80\%  &  7.80\% & 7.80\%  &  8.33\% &  16.17\% \\
          & BlenderBot  &  0.03  & 0.07 & 8.75\%  &  8.75\%  & 8.75\%  & 9.56\%  & 21.01\%  \\          & CTRLStruct w/o Total\_Loss &    0.03   &   0.06    &  7.83\%     &    7.83\%   &  7.83\%     & 8.40\% & 17.54\%\\
          & CTRLStruct w/o WR\_Loss &    \textbf{0.04}   &   \textbf{0.08}    &    9.27\%   &     9.27\%    &     9.27\%    & 10.13\%  & 24.32\%\\
          & CTRLStruct &    \textbf{0.04}   &   \textbf{0.08}    &   \textbf{9.52\%}    &   \textbf{9.52\%}    &   \textbf{9.52\%}    &  \textbf{10.46\%} & \textbf{26.54\%}\\
    \midrule
    \multirow{8}{*}{DailyDialog} & Seq2Seq &    0.09  &   0.07    &    9.43\%   &   9.43\%    &   9.43\%    & 13.00\% & 18.49\%\\
          & CVAE  &    0.09   &    0.06   &   9.22\%    &    9.22\%    &   9.22\%     &  12.49\% &17.97\%\\
          & BART  &   0.14  &  0.12   &   16.38\%    &    16.38\%   &     16.38\%  & 19.86\% & 27.81\%\\
          & DialoGPT & 0.14 & 0.11 & 13.22\% & 13.22\% & 13.22\% & 16.41\% & 23.14\% \\
          & BlenderBot &  0.16 &  0.14 & 17.89\% & 17.89\% & 17.89\% & 22.05\% & 31.14\%\\
     & CTRLStruct w/o Total\_Loss &   0.14    &    0.12   &   15.81\%   &   15.81\%    &  15.81\%     & 19.32\% &  26.62\%\\
          & CTRLStruct w/o WR\_Loss &   0.19    &   \textbf{0.18}    &    22.86 \%   &   22.86\%    &  22.86\%     & 27.27\% &  37.08\%\\
          & CTRLStruct &    \textbf{0.20}   &   \textbf{0.18}    &   \textbf{23.05\%}    &   \textbf{23.05\%}    &   \textbf{23.05\%}  & \textbf{27.78\%} & \textbf{37.70\%} \\
    \bottomrule
    \end{tabular}
 }
  \label{tab:addlabel11}
\end{table*}

\section{Results and Analysis}
\subsection{Dialogue Response Generation (RQ1)}
As suggested in the empirical study of dialogue evaluation \cite{DBLP:conf/emnlp/LiuLSNCP16}, some automatic metrics are originally built for machine translation and can't systematically evaluate the quality of generated dialogue. So we combine automatic evaluation with human evaluation. Experimental results are listed in Table \ref{tab:gen}. In automatic evaluation, CTRLStruct gets the highest BLEU-1/2 score in DailyDialog and the highest BLEU-2 score in PersonaChat where DialoGPT scores better on BLEU-1. Our model remains ahead in Distinct-1/2 in DailyDialog dataset, but fails to surpass DialoGPT in PersonaChat dataset. In ROUGE-L, BlenderBot performs the best in both datasets.

As to human evaluation, we invited fifty people to communicate with eight models trained on different datasets in double blind experiment. Notice that evaluators know which models are trained on the same dataset for fair comparison. Results show that all the models have the ability to generate fluent dialogue without obvious grammatical or spelling errors. Superiority of CTRLStruct is reflected in coherence evaluation, where the best score is more than one point higher than the lowest score. During informativeness evaluation, DialoGPT performs better in PersonaChat dataset, but in DailyDialog CTRLStruct is able to generate complicated utterances from vocabulary and structure perspective. In comprehensive assessment, CTRLStruct gets the highest score in both datasets. All the experiments indicate that CTRLStruct performs better overall than several strong dialogue generation models.

\subsection{Topic Control Quality (RQ2)}
In this part, we test whether the generated responses of different models are related to their supposed topics and how well dialogue structure control the topics in generation process. We view the ground truth answer's topic in test set as supposed topic and turn the evaluation into a multi-classification problem, where the generated response's topic needs to be correctly classified to its ground truth response's topic. Since the true labels of topics are unable to know, pseudo-labels, namely identity numbers of topic clusters, are utilized in evaluation. These pseudo-labels are obtained through the clustering process in dialogue structure modeling part of CTRLStruct. The labels of generated responses are also calculated through the dialogue structure modeling part in CTRLStruct for fair comparison. So the multi-classification problem is formulated as given $k$ categories ($k$ is the total number of clusters) and some samples (generated responses), we need to assess if the model has the ability to perform right topic classifications.

Table \ref{tab:addlabel11} shows the experimental results of evaluating different models' response relatedness to topics in DailyDialog and PersonaChat. CTRLStruct outperforms its counterparts in macro-$\emph{F}1$, micro-$\emph{F}1$, Hard Topic Hit Accuracy (HTHA) and Soft Topic Hit Accuracy (STHA) on both datasets, indicating that our method can generate more topic-related responses compared to other baselines. When the similarity threshold $\varphi \geq 0.90$, HTHA and STHA have the same results, indicating the topic segmentation is good. With the similarity constraint becoming loose, the topic hit accuracy surges. Compared to the results in DailyDialog, scores in PersonaChat is rather low. We attribute this phenomenon to the low quality of PersonaChat dataset. PersonaChat isn't consisted of human conversations like DailyDialog and its topic diversity is low. So when we preset more topic categories than it actually has, a lot of noise will be introduced and damage CTRLStruct.

\subsection{Utterance Representation (RQ3)}

\begin{table}[t]
  \centering
  \caption{Results showing generalization ability of CTRLStruct on GPT2 in two datasets.}
    \resizebox{\columnwidth}{!}
    {
    \begin{tabular}{ccccc}
    \toprule
    \textbf{Dataset} & \textbf{Model} & \textbf{HTHA}  &  \textbf{STHA}(0.85) &  \textbf{STHA}(0.80)\\
    \midrule
    \multirow{2}{*}{PersonaChat} 
          & GPT2 &   5.72\%  &   6.04\%  &   12.02\% \\
          & GPT2 + CTRLStruct  &   \textbf{6.84\%}    &    \textbf{7.28\%}  &   \textbf{14.88\%}  \\
    \midrule
    \multirow{2}{*}{DailyDialog} 
          & GPT2 &   11.73\%    &   14.30\%  &   20.59\%  \\
          & GPT2 + CTRLStruct  &   \textbf{16.95\%}  &   \textbf{20.70\%} &    \textbf{29.13\%} \\
    \bottomrule
    \end{tabular}
 }
  \label{tab:rq4}

\end{table}

\begin{figure}
\centering
\includegraphics[width=\linewidth]{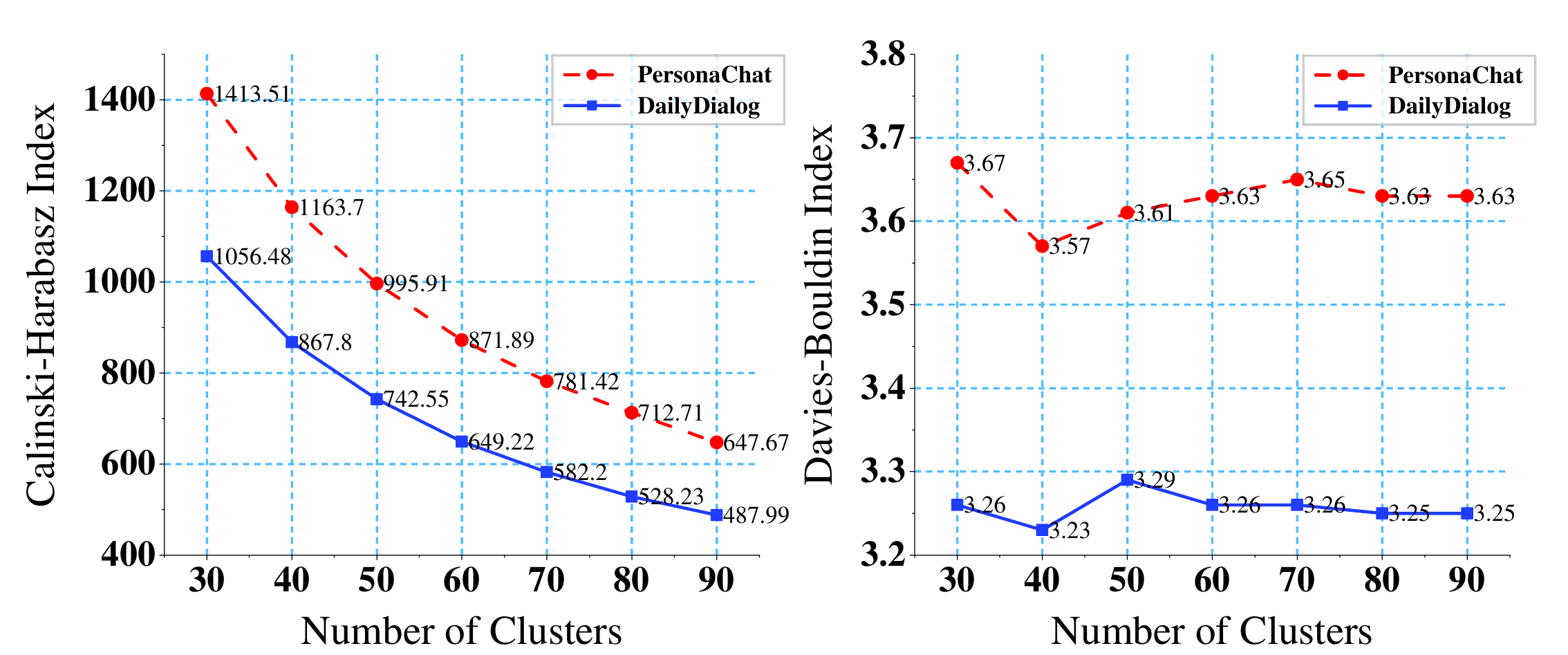} 
\caption{Line charts showing the influence of different cluster numbers on clustering performance.}
\Description[Line charts of evaluating the effect of cluster numbers]{Line charts showing the influence of different cluster numbers on clustering performance of CTRLStruct in DailyDialog and PersonaChat.}
\label{kmeans}
\end{figure}

Under the assumptions that utterances with similar meanings are supposed to get closer in the representation space, we evaluate utterance representation quality of different models through performing clustering on utterance representations and analyze the metrics of clusters. As shown in Table \ref{tab:2}, CTRLStruct outperforms other methods in both Calinski-Harabasz Index and Davies-Bouldin Index on two datasets. BERT and SimCSE score similarly in DailyDialog, but the latter gets a poor performance in PersonaChat.

Moreover, we test the influence of cluster numbers to CTRLStruct. As shown in Figure \ref{kmeans}, with the number of clusters increasing from $30$ to $90$, Calinski-Harabasz Index drops fast first and then slow. As to Davies-Bouldin Index, the value fluctuates when the number of clusters increases in both datasets. One more premise to consider is that the finer topic segmentation is, the higher quality dialogue structure modeling will achieve. A small number of topics will affect the control ability of dialogue structure. So it's supposed to increase the number of clusters as much as possible without damaging the clustering quality. Combining all the factors, we draw the conclusion that in DailyDialog the best $K$ number in K-Means is $60$ and in PersonaChat the best $K$ number is $50$. Generally speaking, CTRLStruct is robust to the number of clusters.

\subsection{Generalization Ability (RQ4)}
CTRLStruct is a framework of discovering dialogue structure from unlabelled corpus and conducting controlled generation, so it's agnostic to the type of backbone models. Original CTRLStruct utilizes BART which is encoder-decoder architecture as backbone. We conduct experiments on GPT2 \cite{radford2019language} which belongs to decoder-only architecture to see whether CTRLStruct can improve its coherence. Results are listed in Table \ref{tab:rq4}. The topic hit rate of GPT2 increases when integrated with CTRLStruct. However it should be noted the proposed dialogue structure is generated through Transformer encoder, so it can’t be applied to none-Transformer network.

\subsection{Ablation Study}
We conduct ablation study on CTRLStruct from three perspectives: utterance representation, response generation, and dialogue structure control. As illustrated in Table \ref{tab:2}, BERT outputs original representations and can be viewed as CTRLStruct without total loss. SimCSE is equivalent to CTRLStruct without Relative Correlation Loss. We notice that utterance representation performance is gradually improved with the constraint of Absolute Correlation Loss and Relative Correlation Loss. In generation stage, we assess CTRLStruct without Weak Relativity Loss and CTRLStruct without total loss. Results in Table \ref{tab:gen} show that CTRLStruct performs better than that without Weak Relativity Loss, and CTRLStruct without total loss even performs worse than BART. In dialogue structure control stage, Figure \ref{tab:addlabel11} shows similar results. The constraint of Weak Relativity Loss can slightly improve CTRLStruct's topic hit accuracy. But CTRLStruct without Total Loss gets poor performance on all indicators. We believe wrong dialogue structure caused by poor utterance representation can mislead response generation and cause bad performance in both topic control and response generation, which is consistent with previous opinion that good utterance representation leads to better dialogue structure and good dialogue structure leads to better generated dialogues.

\begin{table}[t]
  \centering
  \caption{Results of clustering performance on utterance representation, where CHI and DBI stand for Calinski-Harabasz Index and Davies-Bouldin Index respectively. Weak Relativity Loss is abbreviated to WR$\_$Loss.}
\resizebox{\columnwidth}{!}
    {
    \begin{tabular}{ccccc}
    \toprule
    \multirow{2}{*}{\textbf{Model}} & \multicolumn{2}{c}{\textbf{DailyDialog}} & \multicolumn{2}{c}{\textbf{PersonaChat}} \\
         & \multicolumn{1}{c}{\textbf{CHI} $\uparrow$} & \multicolumn{1}{c}{\textbf{DBI}$\downarrow$} & \multicolumn{1}{c}{\textbf{CHI} $\uparrow$} & \multicolumn{1}{c}{\textbf{DBI} $\downarrow$} \\
    \midrule
    BERT  &   566.79    &   3.62    &    848.64   &  3.83\\
    SimCSE &   566.64  &    3.57   &    386.52   &  4.36\\
    CTRLStruct w/o WR\_Loss &   618.42    &   3.28    &    857.78   &  3.65\\
    CTRLStruct &  \textbf{649.22}    &    \textbf{3.26}   &    \textbf{871.89}   &  \textbf{3.63}\\
    \bottomrule
    \end{tabular}
    }
    \label{tab:2}
\end{table}

\subsection{Case Analysis}
We analyze some cases in response generation and dialogue structure modeling, including communications with CTRLStruct to show its advantages and utterances randomly selected in different topic clusters. Due to space limitations, details are shown in Appendix.

\section{Conclusion}
In this paper, we present a novel framework CTRLStruct, which combines contrastive learning, clustering, and imitation learning to effectively capture dialogue utterance representation and construct topic-level dialogue structure. Moreover, dialogue structure which helps track topic flow in conversations is integrated into open-domain response generation. Experiments confirm the superiority of CTRLStruct to other strong dialogue generation models in generating coherent and topic-related conversations.

\begin{acks}
This research is supported by the National Natural Science Foundation of China (No.62106105), the CCF-Tencent Open Research Fund (No.RAGR20220122), the Scientific Research Starting Foundation of Nanjing University of Aeronautics and Astronautics (No.YQR21022), and the High Performance Computing Platform of Nanjing University of Aeronautics and Astronautics.
\end{acks}

\bibliographystyle{ACM-Reference-Format}
\bibliography{ref}
\clearpage

\appendix

\begin{figure*}
\centering
\includegraphics[width=\linewidth]{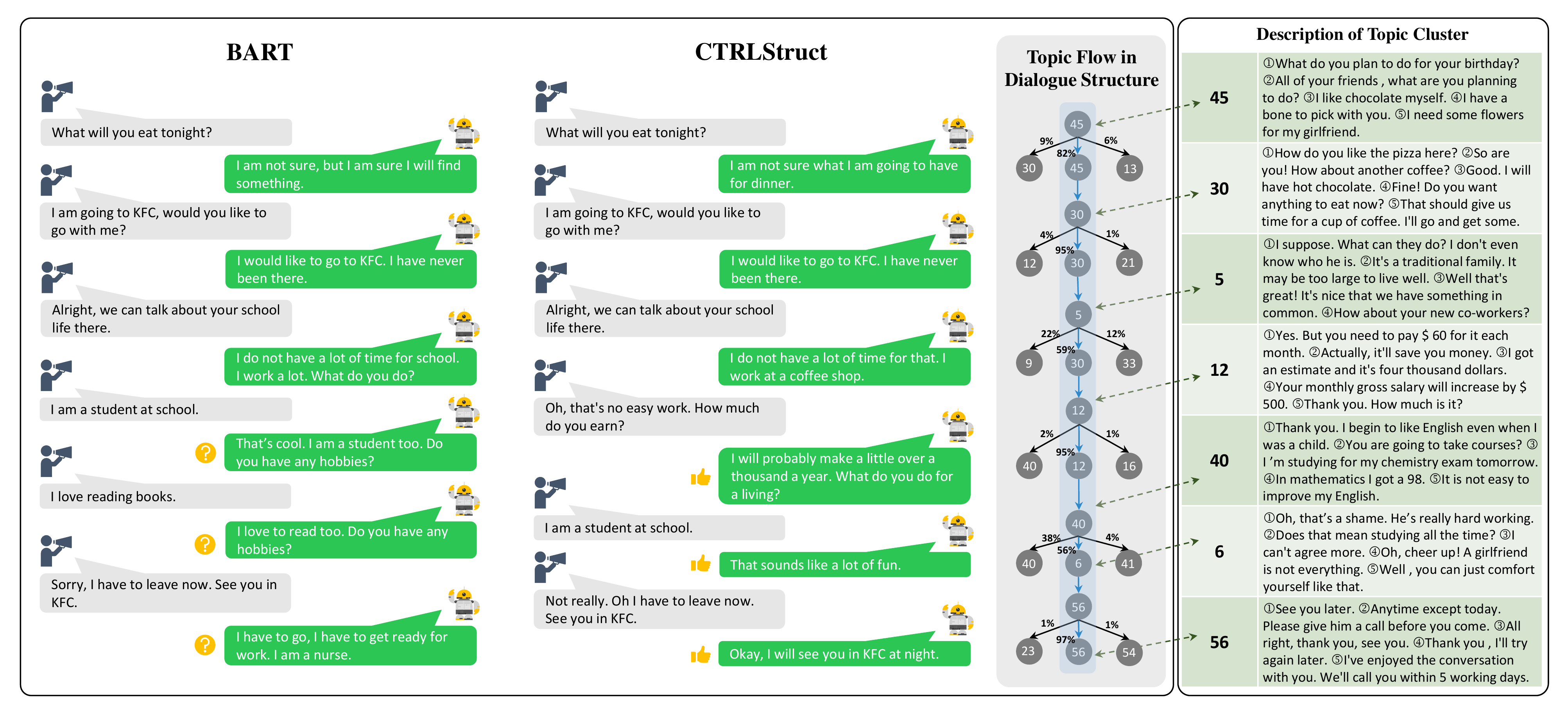}
\caption{Dialogue history between a person A and a chitchat robot B where responses with thumbs up emoji show superiority of CTRLStruct to BART. Topic flow in dialogue structure with topic transition probability is shown on the right.}
\Description[Case study of CTRLStruct compared to BART]{The picture on the left shows dialogue history of BART and CTRLStruct while the picture on the right shows description of the selected topic cluster.}
\label{apd}
\end{figure*}

\section{Case Study}
In case study, we present more details about response generation and utterance representation and analyze these cases.
\subsection{Response Generation}
To further dissect the ability of CTRLStruct in dialogue response generation, we communicate with BART model and CTRLStruct model respectively and analyze the dialogue history. BART is chosen for comparison because CTRLStruct can be viewed as dialogue structure enhanced BART. The differences between generated dialogues are able to prove the value of dialogue structure modeling. Conversations shown in Figure \ref{apd} happened between a person A and a chitchat robot B, which has several topic-level transitions, including A asking B for dinner together, A talking about school life at dinner, A inquiring B's job and A saying goodbye to B. Utterances with thumbs-up emoji and question mark emoji show the differences between two models. We notice that there exists inconsistencies in BART model like the contradictory between ``I am a nurse.'' and ``I am a student, too.''. Also, BART keeps asking the same questions ``Do you have any hobbies?''. Moreover, BART can't give satisfying response when human say ``See you in KFC.''. We attribute these to model's insensibility to topic transitions. Such problems are greatly eased in CTRLStruct which is topic-aware in multi-turn conversations. CTRLStruct can quickly realize the topic transitions from B's job to A's work and when person A has to leave, our model reacts properly instead of making unrelated statements.

\begin{figure}
\centering
\includegraphics[width=\linewidth]{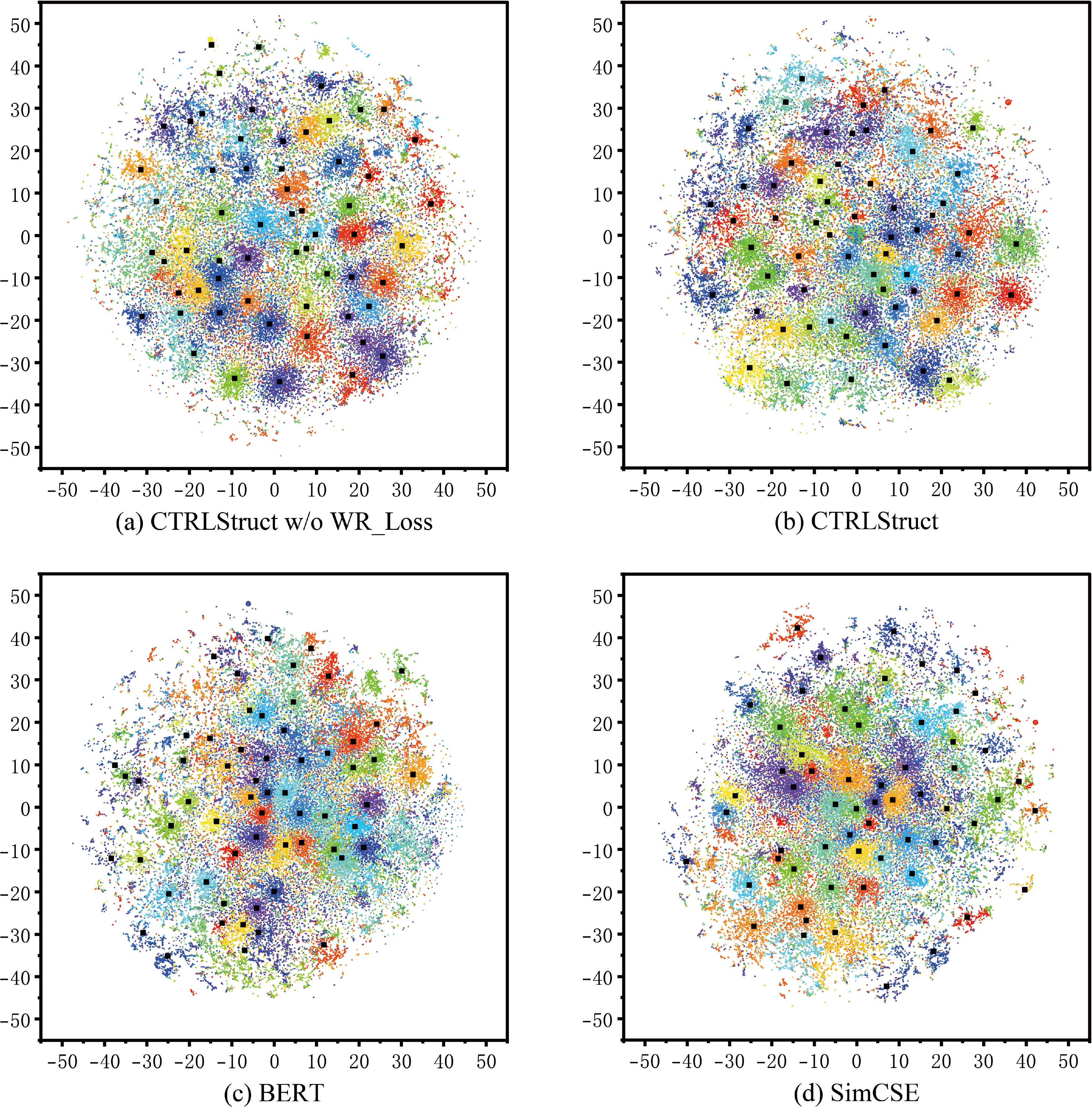}
\caption{Scatter diagrams of different models by t-SNE in DailyDialog dataset.}
\Description[Four pictures of visualizing topic clusters]{We plot scatter diagrams of four different methods to vividly show the utterance representation quality. The cluster number is set as sixty.}
\label{f2}
\end{figure}

We discover the topic transition graph of the given dialogue history and present it in Figure \ref{apd}. The trajectory marked blue is the chosen topic flow generated by CTRLStruct, each selected topic has the highest transition probability at its stage. Topics besides the chosen topic in topic flow graph are the second and third possible options. Descriptions of each topic cluster are on the right and marked green. The descriptions are shown in the form of randomly selecting several utterances in corresponding clusters. For each cluster we select five different utterances. Descriptions are basically consistent with generated responses, which shows CTRLStruct can model high quality dialogue structure and leverage the predicted topic information for response generation.

\begin{table*}
  \centering
  \caption{Examples of utterances randomly selected from clusters formed by CTRLStruct in PersonaChat and DailyDialog.}
  \resizebox{\textwidth}{!}
  {
    \begin{tabular}{cll}
    \toprule
    \textbf{Cluster}
          & \multicolumn{1}{c}{\textbf{PersonaChat}} & \multicolumn{1}{c}{\textbf{DailyDialog}} \\
    \midrule
    \multirow{6}{*}{\textbf{a}} &I bet that is a nice workout. I have 2 older brothers, how many siblings do you have?       & In what business are you most interested in this organization? \\
          & I am sorry to hear that. I have two brothers and two sisters older than me.      & Have you had any experience with sales work? \\
          & Probably longer than us. I wish one of my four sisters was a mermaid!      & There are computers at our EDD offices for you to use in your job search. \\
         &Me too! We have a second child on the way. I am pregnant now.       &  I'd like to start at 3000 yuan a month.\\
          & Me too. Best friend does not even come close. Maybe brother from another mother?      &  I've worked here for two years. And I want to do something different.\\
          & Oh that is great. My girlfriend does all the cooking. Do you have a big family?      &  How would you describe your relationship with our boss?\\
    \midrule
    \multirow{6}{*}{\textbf{b}} &I live in providence and work at a publishing building downtown.       &  I had a terrible quarrel with my neighbor yesterday.\\
          & Hello, how are you? I am well now that i am not a slave.      &  I'm having a hard time getting the information.\\
          & Hey I am jeff and my favorite sport is soccer, how are you?      &  As far as I know, insomnia is usually caused by stress. Are you stressed at all?\\
          & Hi, how are you? I own a small business. I help clients win money.      &  I just lost my job. My boss just told me.\\
          & Oh, I love gardening, although I am color blind and cannot see the flowers very well!      &  We're having some friends over. What are you doing?\\
          & I am a professional dancer and I like listening to music that is not country.      &  You don't look too well. What's going on?\\
    \midrule
    \multirow{6}{*}{\textbf{c}} &Yeah thanks you are so nice! I will give you fresh eggs in return.       &  I have no idea where to get it from.\\
          & Maybe we can meet someday. There is an amazing taco truck in town.      &  What would you like to have?\\
          & Are you vegetarian? I love cooking, it would be neat to fix a vegetarian meal.      & You want to go get a facial with me today? \\
          &That is awesome, that would work great for me, since I am highly educated.       &  What kind of information would you like?\\          & I make good smoothies! I am working at a smoothie shop while i attend college.      &  You don't know where you want to look for one?\\
          &I heard games can help too. I like drawing and I help them learn to draw.       &  Do you know what kind you want?\\
    \midrule
    \multirow{6}{*}{\textbf{d}} &That is kind of superficial of you, but that is why I like you sherry!       &  Hi, my name is Lean, and I'm from Russia.\\
          & A little too violent for me. I mostly spend my time hiking with my dog.      &  Hello, Mrs. Taylor. I'm here to pick up Diane.\\
          & Oh wow, haha. Right into it, eh? I love my life goals. I feel good about them.      & Hello, Rachel. It's glad to see you here. \\
          & I do not mind swimming alone, but I need someone to go with me to weekend concerts.      &She’s very well, too, Helen. Goodbye, Helen. Nice to see you.  \\
          & Oh that is terrible! I understand that too. Its very real. I am opening a grocery store.      &  Yes, this is Jin Lili. What can I do for you?\\
          & I would love to live there! My dad cannot find work there though, he is a cop.      & Good morning, Ms Chan. What can I get you today? \\
    \bottomrule
    \end{tabular}
}
  \label{tab:cluster}
\end{table*}

\subsection{Utterance Representation}
Table \ref{tab:cluster} presents some utterance samples in topic clusters to vividly show CTRLStruct's ability of gathering semantically similar utterances. Clusters and utterances in clusters are all randomly selected. In PersonaChat dataset, clustering performance is not as good as it supposed to be. Cluster \textbf{a} is about family business. Cluster \textbf{b} can be defined as hobbies, but topic about jobs and work is probably mixed in. The quality of cluster \textbf{c} and cluster \textbf{d} is poor and it's hard to define them as one specific topic. We blame it to the low quality of PersonaChat as an open-domain dialogue dataset. Utterance in PersonaChat usually contains too many topics of different aspects and is not appropriate as response from human perspective. The noisy semantic information of utterances will confuse model and degrade performance.

The quality of DailyDialog is much higher than PersonaChat, so the clustering performance in DailyDialog dataset is better correspondingly. In cluster \textbf{a}, all the selected utterances are about jobs, including job hunting, salary, relationship and so on. Cluster \textbf{b} is about concerns for others. Cluster \textbf{c} contains different kinds of queries. Cluster \textbf{d} is about greetings to others. From the selected utterances in DailyDialog, we suppose CTRLStruct successfully captures topic-level meanings without labelled information and performs high-quality clustering.

We also plot scatter diagram of clusters to show different model's performance in DailyDialog dataset. These diagrams are plotted through t-SNE \cite{van2008visualizing} which reduces the dimension of original vectors. Since the clustering task of gathering utterances with similar topics is complex, the effect isn't that significant under dimensional reduction. But we can still notice in Figure \ref{f2}, large numbers of outliers surround the clusters in BERT and SimCSE. Some samples in cluster can't even gather together and form a shape. CTRLStruct performs best in both stabilizing cluster shape and eliminating outliers.

\section{Limitations and Future Work}
In this section we will discuss the limitations of our work and possible improvement in the future. Firstly, due to the limited computing resources, we set the batch size in utterance representation learning as $128$. Larger batch size may lead to better representing performance in contrastive learning. Secondly, when utterance representations are obtained, CTRLStruct applies K-Means as clustering methods. Actually several methods including deep clustering and agglomerative clustering are tested and K-Means performs the best with the least computational resources. However K-Means has some obvious drawbacks like the convex and isotropic assumption. We hope to find some clustering methods that are more suitable under this problem settings. Thirdly, in the controlled generation part, simply bringing utterance representation closer to topic cluster center is an unstable control method and can't make the full use of discovered dialogue structure. We believe by designing new decoding strategy or introducing symbolic rule constraint in generation procedure can effectively improve performance. Moreover, there are barely accessible large high quality open-domain dialogue datasets. As the experimental results in this paper illustrate, the model performance is largely associated with dataset quality. If high quality open-domain dialogue datasets are accessible, we can train CTRLStruct and leverage the pre-trained dialogue structure as commonsense knowledge for downstream dialogue generation training in the future.

\end{document}